\documentclass[conference]{IEEEtran}
\IEEEoverridecommandlockouts
\usepackage{cite}
\usepackage{amsmath,amssymb,amsfonts}
\usepackage{algorithmic}
\usepackage{graphicx}
\usepackage{textcomp}
\usepackage{xcolor}
\usepackage{hyperref}
\usepackage{pgfplots}
\usepackage{subfigure}
\usepackage{multirow}
\usepackage{booktabs}
\usepackage[utf8]{inputenc}
\usepackage{tikz} 
\usetikzlibrary{decorations.markings}

\def\BibTeX{{\rm B\kern-.05em{\sc i\kern-.025em b}\kern-.08em
    T\kern-.1667em\lower.7ex\hbox{E}\kern-.125emX}}
    \pgfplotsset{compat=1.17}

\begin{document}

\title{Causal discovery for time series with latent confounders\\
\thanks{I thank Paul-Christian Buerkner for helpful discussions and
suggestions.}
}

\author{

\IEEEauthorblockN{Christian Reiser}
\IEEEauthorblockA{
\textit{University of Stuttgart}\\
Stuttgart, Germany \\
christian.reiser@insightme.org
}





}

\maketitle

\begin{abstract}
Reconstructing the causal relationships behind the phenomena we observe is a fundamental challenge in all areas of science.
Discovering causal relationships through experiments is often infeasible, unethical, or expensive in complex systems. 
However, increases in computational power allow us to process the ever-growing amount of data that modern science generates, which led to an emerging interest in the problem of causal discovery from observational data.
This work evaluates the LPCMCI algorithm, which aims to find generators compatible with a multi-dimensional, highly autocorrelated time series while some variables are unobserved.
We find that LPCMCI performs much better than a random algorithm that mimics not knowing anything but is still far away from an optimal detection.
Furthermore, LPCMCI performs best on auto-dependencies, then contemporaneous dependencies, and struggles most with lagged dependencies.
The source code of this project is available online\footnote{Sourcecode: \href{https://github.com/christianreiser/correlate/blob/master/causal_discovery/LPCMCI/compute_experiments.py}{https://github.com/christianreiser/correlate/blob/master/causal\_discovery/LPCMCI/compute\_experiments.py}}.
\end{abstract}


\section{Introduction}
Reconstructing the causal relationships behind the phenomena we observe is a fundamental challenge in all areas of science. It helps to build physical models and predict the effect of interventions\cite{pearl_book_2018}.
The goal is to distinguish direct and indirect dependencies between variables, determine the dependencies' directionality, and identify common drivers.
The standard approach is to learn causal relationships through conducting experiments. However, interventions are often infeasible, unethical, or expensive in complex systems such as earth system science or healthcare\cite{mcelreath_statistical_2018}.
Meanwhile, modern science generates a growing amount of data, which we can process due to increases in computational power.
The availability of data and abundance of computing power to process it has led to an emerging interest in the problem of causal discovery from observational data\cite{runge_causal_2018}.

In our work, we generate highly auto-correlated multi-dimensional time-series with latent variables that have similar properties to a dataset used in a study for mood prediction\cite{reiser_predicting_2022}. We briefly review methods that aim to reconstruct generators compatible with the observed data and elaborate on and evaluate a conditional independence constraint-based algorithm called LPCMCI\cite{gerhardus_high-recall_2021} in more detail.


\section{Problem description and notation}
\subsection{Preliminaries}
Consider a multivariate state generator that we want to reconstruct from data.
We assume it follows a vector-auto-regressive process\footnote{Auto-regressive process is a model, where the output variable depends only on its own lagged values and on a stochastic noise term. A \textit{vector-}auto-regressive process extends the auto-regressive model to multiple dimensions. As explained later, the output of our vector-auto-regressive process additionally depends on its current values.}, similar as used in \cite{gerhardus_high-recall_2021}, and described by the structural causal model (SCM)
\begin{equation}
V_{t}^{j}=f_{j}\left(pa\left(V_{t}^{j}\right), \eta_{t}^{j}\right) \quad \text { with } j=1, \ldots, \tilde{N},
\end{equation}
generating a multivariate time series $\mathbf{V}^{j}=\left(V_{t}^{j}, V_{t-1}^{j}, \ldots\right)$ for $j=1, \ldots, \tilde{N}$.
The functions $f_i$ describe mechanisms in the physical world which depend on a set of causal parents $p a\left(V_{t}^{j}\right) \subseteq\left(\mathbf{V}_{t}, \mathbf{V}_{t-1}, \ldots, \mathbf{V}_{t-p_{t s}}\right)$ and jointly independent noise variables $\eta_{t}^{j}$. $p_{t s}$ represents the maximal time lag between cause and effect.

We assume \textit{causal stationary}, which means that dependencies between all pairs of variables $\left(V_{t-\tau}^{i}, V_{t^{}}^{j}\right)$ with the time lag $\tau \geq 0$ is the same as for all time-shifted pairs $\left(V_{t^{\prime}-\tau}^{i}, V_{t^{\prime}}^{j}\right)$\cite{runge_causal_2018}. Therefore, causal stationarity simplifies the problem by reusing a momentary dependency found at one point in the time series and setting its dependency to all other time-shifted pairs.
We allow for \textit{contemporaneous} ($\tau=0$) effects because, in many applications, the measurement frequency is often slower than the time scale of the causal processes. Although we allow for contemporaneous effects exist, we still assume \textit{acyclicity}, which means the causal graph has no cycles.
The SCM has \textit{latent variables}, which are variables in the data-generating SCM that are unobserved. We choose to include latent variables, because many real-world applications often only record a subset $\mathbf{X}=\left\{\mathbf{X}^{1}, \ldots, \mathbf{X}^{N}\right\} \subseteq \mathbf{V}=\left\{\mathbf{V}^{1}, \mathbf{V}^{2}, \ldots\right\}$ of the time series with $N \leq \tilde{N}$. 
However, we assume the absence of \textit{selection variables}, meaning there are no variables determining which measurements to include in or exclude from the data sample. Breaking this assumption can lead to selection bias and invalidate results.
Furthermore, we assume \textit{faithfulness}, meaning that conditional independence (CI) in the observed distribution $P(\bf{X})$ is due to the causal structure of the underlying process instead of mere chance\cite{gerhardus_high-recall_2021}.

\subsection{Implications of latent variables}
\label{sec:implLatent}
This section explains the implications of latent variables, relying on graph theory.
While it briefly explains the graphical terminology, the interested reader can find more information about MAGs and PAGs in \cite[section 9.4.1]{peters_elements_2018}.

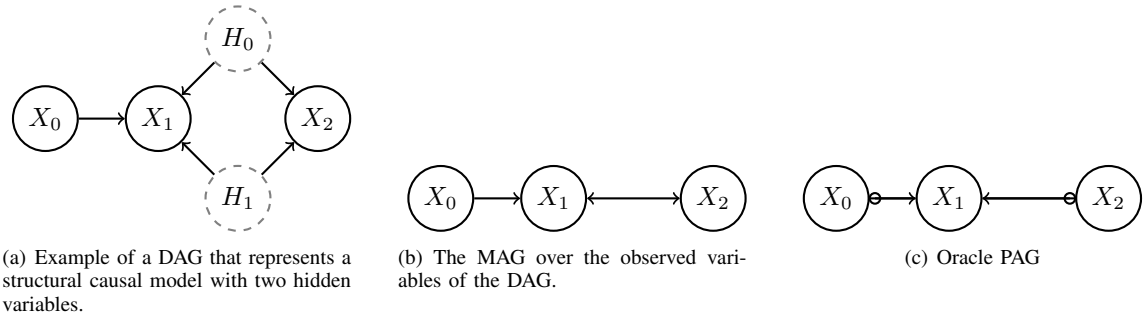
\begin{figure}[htbp]
    \centering
             \subfigure[Example of a DAG that represents a structural causal model with two hidden variables.]{
   \begin{tikzpicture}[node distance={15mm}, thick, main/.style = {draw, circle}, hidden/.style = {draw=gray,dashed, circle}] 
\node[main] (0) {$X_0$}; 
\node[main] (1) [right of=0]{$X_1$}; 
\node[hidden] (2) [above right of=1] {$H_0$}; 
\node[hidden] (3) [below right of=1] {$H_1$}; 
\node[main] (4) [above right of=3] {$X_2$}; 
\draw[->] (0) -- (1); 
\draw[<-] (1) -- (2); 
\draw[<-] (1) -- (3);
\draw [->](2) -- (4); 
\draw [->](3) -- (4); 
\end{tikzpicture}}\hspace{5mm}
    \subfigure[The MAG over the observed variables of the DAG.]{
   \begin{tikzpicture}[node distance={15mm}, thick, main/.style = {draw, circle}, hidden/.style = {draw=gray,dashed, circle}] 
\node[main] (0) {$X_0$}; 
\node[main] (1) [right of=0]{$X_1$}; 
\node[main] (4) [above right of=3] {$X_2$}; 
    \draw[->] (0) -- (1); 
    \draw[<->] (1) -- (4); 
\end{tikzpicture}}\hspace{5mm}
    \subfigure[Oracle PAG]{
   \begin{tikzpicture}[node distance={15mm}, thick, main/.style = {draw, circle}, hidden/.style = {draw=gray,dashed, circle}, o/.style={
        shorten >=#1,
        decoration={
            markings,
            mark={
                at position 1
                with {
                    \draw circle [xshift=-#1,radius=#1];
                }
            }
        },
        postaction=decorate
    },
    o/.default=2pt] 
\node[main] (0) {$X_0$}; 
\node[main] (1) [right of=0]{$X_1$}; 
\node[main] (4) [above right of=3] {$X_2$}; 
    \draw[->] (0) -- (1); 
    \draw[o] (1) -- (0); 
    \draw[o] (1) -- (4); 
    \draw[<-] (1) -- (4); 
\end{tikzpicture}
    }
    \caption{A comparison of three types of graphs that represent information about a structural causal model with latent variables.}
        \label{fig:DAGMAG}
\end{figure}

Data-generating processes are usually represented as graphs with directed edges that contain no directed cycles. Such graphs are called \textit{directed acyclic graphs (DAGs)}. Two nodes are \textit{adjacent} when there is an edge between them, irrespective of the directionality. Figure \ref{fig:DAGMAG}(a) shows a DAG with two latent variables, $H_0$ and $H_1$. Ideally, we would like to reconstruct such DAGs from the data of its observed variables, here, $X_1$ and $X_2$. However, the reconstruction algorithms do not know how many latent variables there are in the original DAG, and we do not restrict their number. Thus, the algorithm would search among an infinite number of DAGs, which would be infeasible.
Instead, as in the example of Figure \ref{fig:DAGMAG}(b), one can represent each DAG with latent variables by a \textit{maximal ancestral graph (MAG)} over the observed variables. The MAG illustrates variables that are conditionally dependent due to latent confounding with bi-directed edges ($\leftrightarrow$)\cite{richardson_ancestral_2002}. 
MAGs are similar to DAGs but have the following differences in semantics. In a MAG, the tail (-) of an arrow, as in A $\rightarrow$ B, says that A is an ancestor of B. An arrowhead ($<,>$), like the one pointing on B, specifies that B is not an ancestor of A. Also, it does not exclude the possibility that A and B might be related due to latent confounding.
Since an arrowhead in a MAG declares the variable next to it as a non-ancestor, bidirectional edges ($\leftrightarrow$) exclude both variables from being the ancestor of the other variable and thus propose latent confounding between the two.

\begin{table}[htbp] 
\begin{center}
    \caption{Chains of both directions and forks cannot be differentiated via conditional independence. However, they can be differentiated from colliders. $X\!\perp\!\!\!\perp Y$ and $X\!\perp\!\!\!\perp Y|Z$ denote independence and conditional independence between the random variables X and Y.}
\begin{tabular}{@{}l|lll|l@{}}
\toprule
Name         & Chain   & Chain   & Fork    & Collider \\ \midrule
    Graph & $X\rightarrow Z \rightarrow Y$ & $X\leftarrow Z \leftarrow Y$ & $X\leftarrow Z \rightarrow Y$ & $X\rightarrow Z \leftarrow Y$ \\
    Independence & $X\not\!\perp\!\!\!\perp Y$    & $X\not\!\perp\!\!\!\perp Y$    & $X\not\!\perp\!\!\!\perp Y$   & $ \bf{X\!\perp\!\!\!\perp Y}  $       \\
    CI           & $X\!\perp\!\!\!\perp Y|Z$   & $X\!\perp\!\!\!\perp Y|Z$   & $X\!\perp\!\!\!\perp Y|Z$  & $\bf{X\not\!\perp\!\!\!\perp Y|Z}$      \\
    Equivalence & \multicolumn{3}{l|}{one Markov equivalence class} & other Markov equivalence class \\ \bottomrule
\end{tabular}
        \label{tab:equiq}
\end{center}
\end{table}
CI-based methods cannot differentiate all MAGs \cite{richardson_ancestral_2002}. Graphs  containing the same CI information are summarized in a class called \textit{Markov equivalent class}. Table \ref{tab:equiq} shows examples of Markov equivalence for cases without latent confounding. The interested reader can find a find a full explanation of Markov equivalence under latent variables in \cite[section 3.6]{richardson_ancestral_2002}. 
Due to Markov equivalence and contemporaneous effects, we can often not reduce the number of possible MAGs to one. Therefore, we aim to find all MAGs that satisfy the CI and time-ordering constraints. All these possible MAGs can be summarized in one graph, represented by a \textit{partial ancestral graph (PAG)}\cite{zhang_causal_2008}.
PAGs have the same semantics as MAGs except that a PAG can also represent uncertainty whether a link has an arrowhead ($<$ or $>$) or tail (-) with a circle-head ($\circ$), as shown in Figure \ref{fig:DAGMAG}(c). For example, the link C $\circ$$\rightarrow$ D says that D is not an ancestor of C, but we do not know if C is an ancestor of D or not. 
This case leaves us with three possibilities. First, there could be a directed path from C to D; second, the dependency could only be due to latent confounding; third, it could be a combination of both. 

It is difficult to find the PAG that satisfies all CI constraints.
Nevertheless, finding this PAG is our goal. We evaluate the algorithm's performance against the PAG that perfectly satisfies all CI information over the observed variables. We call this perfect PAG \textit{oracle PAG}, and it serves as our ground truth.
Figure \ref{fig:DAGMAG}(c) shows the oracle PAG of the previous example. 

To summarize, we aim to find the PAG, which represents all MAGs that satisfy the conditional independence and time-order constraints of the observed variables of the SCM.


\section{Introduction into methods}
\textit{Granger causality} can infer the causal structure of time series. Intuitively we say that X Granger-causes Y when the prediction of Y from its past is improved by accounting for X's past.
Then Y has to be dependent on the past of X given its own past. Formally that is
\begin{equation}
X \text { Granger-causes } Y: \Longleftrightarrow Y_{t} \not\!\perp\!\!\!\perp X_{\operatorname{past}(t)}  \mid Y_{\operatorname{past}(t)},
\end{equation}
where $\not\!\perp\!\!\!\perp$ denotes dependence and $|$ means 'conditioned on'\cite[p.~204]{peters_elements_2018}.
Clive Granger himself extended Granger causality to a multivariate setting\cite{granger_testing_1980}. However, it can be misleading with the presence of instantaneous effects \cite{granger_clive_recent_1988} and requires that all variables are observed\cite{granger_testing_1980}.

\textit{FullCI} determines the absence of a link between one variable $X_{t-\tau}^{i}$ with time lag $\tau$ and another variable $X_{t}^{j}$ by testing for their independence while conditioning on the past of all other variables $\mathbf{X}_{t}^{-} \backslash\left\{X_{t-\tau}^{i}\right\}$. 
Formally that is
$X_{t-\tau}^{i} $
$\!\perp\!\!\!\perp X_{t}^{j} $
$\mid \mathbf{X}_{t}^{-}$
$ \backslash\left\{X_{t-\tau}^{i}\right\}$ \cite{runge_pcmci_2019}.
A problem is that high dimensionality of the conditioning set decreases the detection power of FullCI \cite{runge_pcmci_2019}.
Usually, the researcher decreases dimensionality by restricting the maximal time-lag $\tau_{max}$ through background knowledge or sets it to the largest time-lag with significant unconditional dependence\cite{runge_pcmci_2019}.
It is actually sufficient to condition only on the parents of the two variables $X_{t-\tau}^{i}$ and $X_{t}^{j}$\cite{pearl_causality_2000}, which allows decreasing the dimensionality of the conditioning set further, especially if the true graph is sparse.

The \textit{PC algorithm}, named after its inventors Peter and Clark, is well known to exploit sparsity by iteratively using small conditioning sets\cite{runge_pcmci_2019}. However, when the size of the conditioning set is very limited, chances increase that true parents are not in the conditioning sets and false-positive links tend to remain\cite{runge_pcmci_2019}.

The \textit{PCMCI} algorithm first runs a version of the PC algorithm adapted for time series, which removes most false links very efficiently, with the trade-off that some false positives remain\cite{runge_pcmci_2019}. In the second step, it tests for \textit{momentary conditional independence (MCI)}
\begin{equation}
\mathrm{MCI}: X_{t-\tau}^{i} \!\perp\!\!\!\perp X_{t}^{j} \mid \left[{pa}\left(X_{t}^{j}\right) \backslash\left\{X_{t-\tau}^{i}\right\}, {pa}\left(X_{t-\tau}^{i}\right)\right]
\end{equation}
to remove the remaining false links\cite{runge_pcmci_2019}.
This resulted in faster runtime and increased detection power than FullCI\cite{runge_pcmci_2019}.
PCMCI allows for multivariate and highly auto-dependent time series. Its extension PCMCI+ also handles contemporaneous links. However, PCMCI and PCMCI+ both assume the absence of latent confounders.

An algorithm that allows for latent confounders is the \textit{Fast Causal Inference algorithm (FCI)}\cite{spirtes_causation_2000}. It is constraint-based and outputs a PAG which represents the possible MAGs.
The tsFCI algorithm adapts FCI to time-series \cite{entner_causal_2010}. It extends FCI by additionally exploiting time-ordering. Furthermore, it assumes stationarity which allows applying a found momentary dependency between a pair of variables to be the same for all other shifted pairs.

So war we discussed methods that restrict the set of possible graphs via conditional independence and time ordering. As mentioned in section \ref{sec:implLatent}, the major downside of conditional independence-based methods is that they cannot differentiate Markov equivalent graphs. 
However, it is possible to differentiate Markov equivalent graphs if they represent \textit{Linear Non-Gaussian Acyclic Models (LiNGAMs)}.
In the bivariate case, if the true SCM is 
\begin{equation}
Y=\alpha X+N_{Y}, \quad N_{Y} \!\perp\!\!\!\perp X
\end{equation}
with continuous random variables $X, Y$, and $N_Y$, then there exist no $\beta\in\mathbb{R}$ and random variable $N_X$ for the SCM in the reverse direction
\begin{equation}
X=\beta Y+N_{X}, \quad N_{X} \!\perp\!\!\!\perp Y
\end{equation}\cite[p.~48]{peters_elements_2018}.
The reason is that the LiNGAMs of both causal directions would be two dependent linear combinations of independent non-gaussian distributed random variables with non-zero coefficients. But the contrapositive of the Darmois–Skitovich theorem\cite{skitovich_linear_1954} says that such linear combinations cannot exist.
Bi-variate LiNGAM algorithms test in which direction the requirement of independent noise is broken ($N_{Y} \!\perp\!\!\!\perp X$ or $N_{X} \!\perp\!\!\!\perp Y$), by regressing the data in both directions linearly. Residuals of the causal direction will be independent, and the residuals in the anti-causal direction will depend on the input variable. 
The interested reader can find further details and visualizations in \cite[p. ~108]{neal_introduction_2020}, a mathematical proof for the bi-variate case in\cite[Appendix C.1]{peters_elements_2018}, and a proof for the multivariate case in \cite{shimizu_linear_2006}.

TS-LiNGAM is an algorithm based on LiNGAM but adapted for time series\cite{hyvarinen_causal_2008}. Like LiNGAM, it assumes linear dependencies and non-Gaussian noise. Since it can identify structure within the Markov equivalence class without relying only on time-ordering, it may have an advantage in handling contemporaneous links.
Limitations of TS-LiNGAM are the restriction to linear relationships, non-Gaussian noise, and instantaneous effects\cite{peters_causal_2013}. Furthermore, Peters et al. show that TS-LiNGAM fails when there is latent confounding, as it wrongly predicts directed edges instead of bi-directed edges\cite{peters_causal_2013}.

\section{Detection power and LPCMCI}
The main challenge of constraint-based methods is increasing detection power, which quantifies the probability of finding a true link between two nodes\cite{runge_pcmci_2019}.
The detection power of conditional independence based methods can be increased by
\begin{itemize}
    \item increasing the number of samples in the dataset \cite{gerhardus_high-recall_2021}, which results in more reliable CI tests,
    \item decreasing the dimensionality of the problem\cite{runge_pcmci_2019},
    \item increasing the causal effect size between the two variables\cite{gerhardus_high-recall_2021},
    \item increasing the statistical significance level of the conditional independence test\cite{gerhardus_high-recall_2021}, and
    \item decreasing the size of the conditioning set until it consists only of the parents of the two variables\cite{runge_pcmci_2019}.
\end{itemize}

In real-world applications, it is often possible to take more samples, decrease the problem's dimensionality by excluding irrelevant variables through expert knowledge, and increase the effect size by enhancing the signal-to-noise ratio of the measurements.
However, these properties are of less interest when developing methods and are usually fixed through the dataset.

Typically, the researcher allows a specific false-positive rate that determines the significance level \cite{gerhardus_high-recall_2021}.
However, one can relax the statistical significance level of the conditional independence tests, which usually increases detection power.
Nevertheless, the downside is that it increases the risk of detecting more false links.
Furthermore, these false links in later steps can also decrease detection power. The reason is that later CI tests of true links might condition on these false links and reduce the CI scores of the true links. Each score that falls below the nominal significance level $\alpha$ leads to the false removal of a true link. More intuitively, false links can "explain away" the effect of true links and thus remove them.
In our test case, we are less interested in keeping the false positive rate strictly below a specific value due to the cost of much higher false-negative rates. We rather optimize alpha to minimize true positives and true negatives equally strong. More details follow in chapter \ref{sec:eval}.

The \textit{LPCMCI} algorithm (for Latent PCMCI) improves its performance by decreasing the size of the conditioning sets by discarding conditioning sets containing known non-ancestors of the two variables of interest but including their known parents\cite{gerhardus_high-recall_2021}. It is challenging to do so because the conditioning sets have to contain the parents of the two variables before the CI-tests are completed. This order is not the case for predecessor algorithms, which first conduct the CI test and then orient the links. Only the orientation phase yields parenthoods.
To overcome the challenge that ancestries and parentships are unknown, LPCMCI entangles CI tests with edge orientation to identify parentships early on. Each time the algorithm gains new knowledge about parent- or ancestorship, it updates its conditioning sets. After the algorithm converges once, it re-initializes the graph but keeps the identified parentships. 
It applies the same process again but extends the conditioning sets with the previously identified parents.
This procedure should remove most spurious links during the second run and assign edgemarks to the remaining links. 
However, the resulting graph still tends to contain false links between variables that are dependent due to latent confounding. 
To remove these false links, the algorithm applies one more round of CI test and edge orientations with a modified rule to select conditioning sets\footnote{For more information on these modified conditioning sets, read about the $napds_t$ sets in definition S5 of\cite{gerhardus_high-recall_2021}.} 
to identify latently confounded links\cite{gerhardus_high-recall_2021}.

The predicted PAG often has Markov equivalent MAGs that cannot be told apart from conditional independence information or time-ordering. They might, however, be discerned through \textit{background knowledge}. For example, when a researcher has prior knowledge that a variable has no effect on another variable or knows the directionality of the effect, the adjacency could be manually removed or oriented before the first CI tests start. 
In addition to the immediate effect of the manual removal or orientation, there is a beneficial indirect effect if it provides new information about parentships or non-ancestorships to improve the conditioning sets.
Although we do not use background knowledge in our experiments, it is desirable to do so because it may lead to more unambiguous edge removal and edge orientations and hence more causal inferences about the underlying data generating process. 



\section{Experiments}
The motivation of our work is to discover the causal dependencies in a real-world dataset about a person's mood and measurements from several consumer services and devices (e.g., sleep, nutrition, weather)\cite{reiser_predicting_2022} with the LPCMCI algorithm. 
However, it is difficult to evaluate against this real-world dataset as the ground truth of its causal dependencies is unknown. Therefore, we apply the LPCMCI algorithm on simulated datasets with similar properties and known ground truth.

\subsection{Data-generation}
The synthetic data generating SCMs are given by
\begin{equation}
V_{t}^{j}=a_{j} V_{t-1}^{j}+\sum_{i} c_{i} f_{i}\left(V_{t-\tau_{i}}^{i}\right)+\eta_{t}^{j} \quad \text { for } \quad t \in\{1, \ldots, T\}, \quad j \in\{1, \ldots, \tilde{N} \}.
\end{equation}
The model consists of $\tilde{N}=11$ variables $V_t^j$. Three of these variables are selected at random and become latent later on.
Every variable $V_t^j$ has a linear auto-dependency $a_j V_{t-1}^{j}$, with $a_j \sim \mathcal{U}(0.3,0.6)$, meaning that the strength of the auto-dependency is drawn from a uniform distribution with a lower bound of 0.3 and an upper bound of 0.6.
In total there are $11$ randomly chosen variable pairs $(V_t^j,V^i_{t-\tau_i})$ with non-zero linear cross-dependencies $f_i \sim \pm \mathcal{U}(0.2,0.5)$. Otherwise $f_i=0$, meaning that all other variables have no cross-dependency.
60\% of the links are instantaneous, meaning they have a time-lag of $\tau =0$. The other 40\% have a time-lag of $\tau=1$.
The noises $\eta^{j} \sim \mathcal{N}(0,\mathcal{U}[0.5,2])$.
Every SCM generates a discrete-time series of length T = 500. The experiment uses data of 4000 SCM. If a SCM generates a non-stationary time series, it is redrawn. This can happen when feedback loops in the SCM lead to trends or periodic patterns in the values or variance.

\subsection{Example}
\begin{figure}[htbp]
    \centering
        \subfigure[Original DAG of the data-generating process. Variables 0, 3, and 5 are the latent variables that generate data and influence the process but are not observed.]{
    \includegraphics[width=0.30\linewidth]{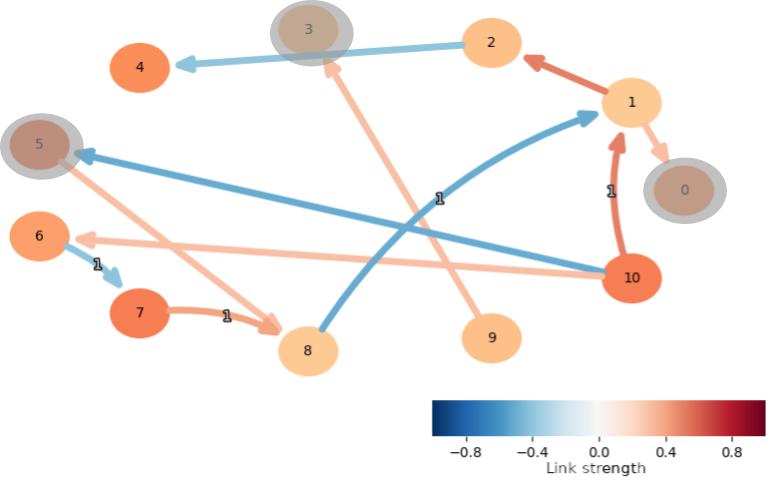}}\hspace{5mm}
             \subfigure[Oracle PAG, which serves as ground truth. It satisfies all conditional independence information of the observed DAG perfectly.]{
             \includegraphics[width=0.30\linewidth]{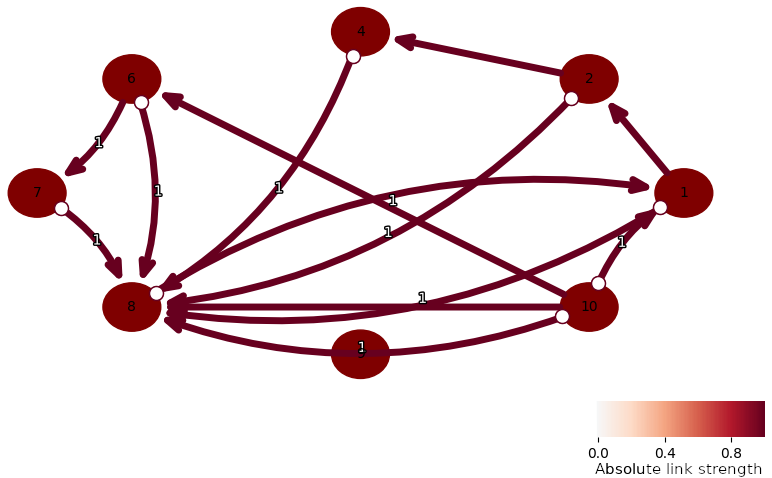}}\hspace{50mm}
    \subfigure[Prediction PAG: The LPCMCI algorithm uses the observable data of the original DAG and returns this predicted PAG, to match the oracle PAG.]{
    \includegraphics[width=0.30\linewidth]{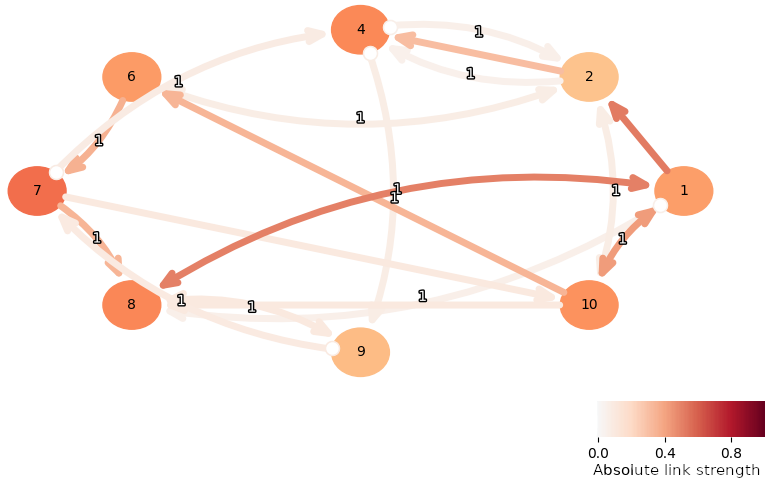}}
    \caption{Comparison of the original data-generating DAG, the oracle PAG, and the predicted PAG of LPCMCI.}
        \label{fig:predicted}
\end{figure}

Figure \ref{fig:predicted}(a) shows the data-generating DAG of $\tilde{N}=11$ auto-dependent variables with eleven cross-dependencies.
The number in the nodes represents the name of the variable.
All auto-dependencies have a time delay of $\tau=1$.
A "1" written on a curved arrow indicates that the cross-dependency has a time-delay $\tau=1$. Straight arrows without a number represent contemporaneous effects. 
The color of an edge or node represents the strength of the cross- or auto-dependency, respectively.

Figure \ref{fig:predicted}(b) shows the oracle PAG, which serves as ground truth. It satisfies all conditional independence information of the observed DAG perfectly.
$\circ$-edgemarks, e.g., as shown in the arrow from node 7 to node 8, represent the possibility that both a tail ("-") or arrowhead ("$<$","$>$") edgemark could exist.
A bidirected edge, e.g., between nodes 1 and 8, indicates that neither of the two variables is a parent of the other, but they are latently confounded.

Figure \ref{fig:predicted}(c) shows the predicted PAG, obtained by the LPCMCI algorithm. The colors represent the absolute link strength. 
It seems like there are many of the false positives have a weak absolute link strength.

\subsection{Evaluation metrics}
\label{sec:eval}
Recall from section \ref{sec:implLatent} that we do not try to reconstruct the DAG that represents the original data-generating SCM. Instead, we aim to find and evaluate against the oracle PAG, which is the PAG that satisfies all conditional independence and time-order constraints of the observed variables of the SCM. 

Our evaluation metrics distinguish between auto-dependencies and cross-dependencies. Furthermore, cross-dependencies are separated into contemporaneous links and lagged links, whereas auto-dependencies always have a lag, as contemporaneous auto-dependencies would violate the assumption of acyclicity.
Furthermore, we distinguish dependencies on whether adjacencies and their correct edgemarks are detected or not. In contrast to the oracle PAG, LPCMCI can additionally return \textit{conflict edgemarks} ("x"), although they will always be evaluated as false positives.
A conflict occurs during edge orientation when orientation rules of two variable triples propose to orient a common edgemark as both tail and head\cite{runge_pcmci_2020}. The underlying problem would be insufficient sample size or at least one assumption violation\cite{runge_tigramite_2022}.

We compute \textit{recall} and \textit{precision} of the adjacency and edge-orientation predictions. Recall (also known as \textit{sensitivity}) indicates the rate of detecting a true link or adjacency. Precision (also known as \textit{positive predictive value}) measures the rate that a predicted link is actually there. Formally, that is $recall = \frac{tp}{tp+fn}$ and $precision = \frac{tp}{tp+fp}$, where
$tp$ stands for true positives, which is the number of edgemarks and adjacencies that are in the oracle PAG as well as the predicted PAG.
$fn$ stands for false negatives,  the number of adjacencies or edgemarks in the oracle PAG but not in the predicted PAG.
$fp$ stands for false positives, the number of adjacencies or edgemarks in the predicted PAG but not in the oracle PAG.
Note that LPCMCI predicts edgemarks for all detected adjacencies and never for undetected adjacencies. It implies that each adjacency-$fn$ leads to two missing edgemarks, which increase the edgemark-$fn$ by two.
Also, vice versa, each adjacency-$fp$ leads to two edgemark predictions that are not in the ground truth, thus increasing the edgemark-$fp$ by two.

A predictor can easily increase precision at the cost of recall, and vice versa. For example, a perfect recall but low precision in adjacency detection can be achieved by blindly proposing all adjacencies to exist. We would not gain any information with such a result, but the arithmetic mean of precision and recall would still be $>0.5$. A more appropriate metric is the \textit{$F_1$-score}
\begin{equation}
F_{1}=2 \cdot \frac{\text { precision } \cdot \text { recall }}{\text { precision }+\text { recall }},
\end{equation}
which is the \textit{harmonic mean} of precision and recall.
To provide intuition about the harmonic mean and the$F_1$-score, let us look at two cases.
\begin{itemize}
    \item Suppose an algorithm does not detect anything in a dataset with positives. It still achieves a precision of 1.00 but at the cost of a recall of 0.00. Then its $F_1$-score would also be 0.00, which describes the situation well as there is no gain in information.
    \item When precision and recall are the same, then the $F_1$-score will also be that number. E.g., when precision and recall are both 1.00, then $F_1$=1.00.
\end{itemize}
In order to reduce the problem to a single-objective optimization problem, we compute the harmonic mean
\begin{equation}
harmonic\_score=\frac{4}{\frac{1}{recall\_adjacencies}+\frac{1}{precision\_adjacencies}+\frac{1}{recall\_edgemarks}+\frac{1}{precision\_edgemarks}},
\label{eq:harmonicscore}
\end{equation}
that aggregates the objectives of precision and recall of edgemarks and adjacencies to a single score.
We also used the $harmonic\_score$ as the objective to optimize the significance level $\alpha=0.26$. More specifically, we decreased the number of SCM realizations from 4000 to 50 to increase speed and searched manually for an $\alpha$ that leads to the best $harmonic\_score$.

To better understand what the LPCMCI algorithm's scores mean, we compare them to the performance of a baseline algorithm that does not have any knowledge and predicts randomly by drawing from a discrete uniform distribution.
In the binary case of adjacencies, the random baseline predicts 50\% of the adjacencies as existent and 50\% as non-existent.
In the case of edge-classification, the random algorithm draws from the set of five classes \{"$\rightarrow$", "$\leftarrow$", "$\leftrightarrow$", "$\circ$$\rightarrow$", "$\leftarrow$$\circ$"\} each with with a chance of 20\%.

Note that if the algorithm predicts the original data-generating DAG, the evaluation would still return a non-perfect score if the predicted DAG differs from the oracle PAG, even though the oracle PAG includes the predicted DAG. More specifically, the DAG over the observed variables of the data-generating SCM is always included in the oracle PAG. But the oracle PAG is often larger, as it includes all MAGs that satisfy the CI and time-order constraints over the observed variables, even if their adjacencies are not in the data-generating DAG. Since the oracle PAG serves as ground truth, every adjacency that is not predicted but in the oracle PAG is evaluated as a false negative, even when not in the data-generating DAG.
For example, Figure \ref{fig:predicted}(b) shows an adjacency from node 6 to 8, which is not in the data-generating DAG of \ref{fig:predicted}(a). Figure \ref{fig:predicted}(c) shows that this adjacency is not detected; thus, the missing adjacency is evaluated as a false negative.

\subsection{Numerical Results}
\begin{table}[!htb]
\begin{center}
        \caption{Numerical results}
        \caption[]{precision, recall, and $F_1$-score of LPCMCI compared to the F1-score of the random baseline algorithm. The table differentiates between edgemarks and adjacencies and further distinguishes between lagged, auto-dependent, and contemporaneous links.}
      \centering
        \begin{tabular}{@{}llllll@{}}
            \toprule
            \textbf{}                  & & precision        & recall  &  $F_1$-score & $F_1$-baseline    \\ \midrule
            \multirow{4}{*}{adjacency} & auto-dependent     & 1.00             & 1.00  &  1.00     & 0.67     \\
                                       & contemporaneous    & 0.60             & 0.77  &  0.67     & 0.29     \\
                                       & lagged             & 0.38             & 0.39  &  0.38     & 0.29     \\
                                       & total              & 0.62             & 0.67  &  0.64     & 0.35     \\ \midrule
            \multirow{4}{*}{edgemark}  & auto-dependent     & 0.74             & 0.73  &  0.73     & 0.26     \\
                                       & contemporaneous    & 0.34             & 0.44  &  0.38     & 0.10     \\
                                       & lagged             & 0.27             & 0.26  &  0.26     & 0.12      \\
                                       & total              & 0.42             & 0.46  &  0.44     & 0.14     \\ 
            \end{tabular}
            \label{tab:precision_recall}
    \end{center}
\end{table}
LPCMCI has an overall performance of $harmonic\_score = 0.53$ on our dataset, while the perfect score would be 1.00. This indicates that LPCMCI has great difficulty predicting the PAGs of our datasets. However, it is important to remind that a random algorithm does not achieve a score of 0.5, which would suggest that LPCMCI is just slightly better than random. Instead, LPCMCI strongly exceeds the performance of the random algorithm, as the random algorithm has only a $harmonic\_score = 0.20$. 

Table \ref{tab:precision_recall} shows the precision, recall, $F_1$-scores of the LPCMCI algorithm, and baseline $F_1$-scores of the different categories, and we discuss these results below.

\subsubsection{Auto-dependent adjacencies}
Within the auto-dependent adjacencies, LPCMCI has a perfect $F_1$-score of 1.0 while the $F_1$-baseline=0.67. Both algorithms score relatively high because there can never be a false positive as all nodes of the oracle PAGs have auto-dependencies. The lack of false positives results in perfect precision for LPCMCI and the random baseline.
In contrast to the baseline, LPCMCI also has perfect recall. This can at least be partially explained by the fact that the performance of LPCMCI grows with stronger auto-correlation\cite{gerhardus_high-recall_2021}, and our dataset has relatively high auto-correlations with values uniformly distributed between 0.3-0.6.

\subsubsection{Auto-dependent edgemark classification}
The edgemark classification of LPCMCI achieves the best results for auto-dependent links with an $F_1$-score of 0.73 while the $F_1$-baseline is only 0.26. One reason for the good performance of LPCMCI is that the edgemark pointing to the node later in time can never be the ancestor of a variable earlier in time. Thus, one edgemark is certainly $>$. 
However, edgemark detection of the variable earlier in time can still fail because it still has the possibilities to be a $<$ (indicating certainty of confounding), $-$ (proposing ancestorship), or $\circ$ (indicating uncertainty of the two).
Another reason for the good performance of LPCMCI is that recall of edgemark detection profits from the high recall of auto-dependent adjacency detection. The reason is that LPCMCI neither had false-negative nor false-positive auto-dependent adjacencies that directly cause false negatives and false positives in edgemark detection. The mechanism of how false adjacency detection causes false edgemark detection was explained in section \ref{sec:eval}.

\subsubsection{Contemporaneous adjacency detection}
Within the contemporaneous adjacencies, LPCMCI has a $F_1$-score of 0.67, which is substantially higher than its $F_1$-baseline=0.29 and higher than the performance of LPCMCI on lagged adjacencies.
One might find it counter-intuitive why LPCMCI performs better on contemporaneous links compared to lagged links, even though lagged links provide additional constraints due to time ordering. However, remember that we evaluate against the oracle PAG, which already takes additional time ordering constraints into account. For example, let us take two links that are not in the SCM but cannot be removed with perfect conditional independence. One of these links is lagged and can be removed due to time-ordering. The other is contemporaneous and cannot be removed with perfect time order information. Then only the contemporaneous link will be in the oracle PAG, which serves as ground truth. As the contemporaneous link is in the ground truth, it will not be evaluated as a false positive. This is one reason why lagged links do not necessarily have a higher evaluation score than contemporaneous links, but it does not explain why the opposite is true.
I still cannot explain why the performance of contemporaneous links is better than on lagged links.
It is, however, also the case in the results of the original LPCMCI paper \cite{gerhardus_high-recall_2021}.

\subsubsection{Contemporaneous edgemark classification}
Within the contemporaneous edgemarks, LPCMCI has a $F_1$-score of 0.38, which is again higher than the performance of LPCMCI on lagged edgemarks. The baseline is $F_1$-baseline=0.10.
A reason for the better performance of LPCMCI on contemporaneous edgemarks compared to lagged edgemarks is that adjacency detection was already better for contemporaneous adjacencies than lagged adjacencies. The effect is the same as for auto-dependencies and is described in section \ref{sec:eval}

\subsubsection{Lagged adjacency detection}
The $F_1$-score of 0.38 of LPCMCI on lagged adjacencies is the lowest score of all adjacency categories but still a little higher than the $F_1$-baseline with 0.29.

\subsubsection{Lagged edgemark classification}
The $F_1$-score of 0.26 of LPCMCI on lagged edgemarks is the lowest score of all adjacency categories but still higher than the $F_1$-baseline with 0.12.
One reason for the low score is the effect of the low performance on lagged adjacencies.

\subsubsection{Conflicts in edgemark proposals}
In our test case, conflicting edgemark proposals occur with a rate of 0.03 and only during the orientation of contemporaneous adjacencies. A higher conflict rate for contemporaneous links seems generally plausible, as time-ordering can solve conflicts for lagged and auto-dependent links but not for contemporaneous links.

\subsubsection{Order independence}
LPCMCI performs best to detect auto-dependencies, then contemporaneous links and struggles most with lagged links, which also happens to be the order in which LPCMCI tests for links. One might thus hypothesize that tests earlier in the order work better than later tests. However, the output of LPCMCI does not depend on the order in which the variables are tested\cite{gerhardus_high-recall_2021}.


\section{Limitations and ideas for future work}
Although LPCMCI ($harmonic\_score=0.53$) strongly outperforms the random baseline ($harmonic\_score=0.20$), it can only partially predict the PAGs of the observed variables of the SCMs, because a perfect score would be $harmonic\_score=1.00$.

A general limitation of LPCMCI is that a measurement frequency lower than the causal processes could violate the assumption of acyclicity of the graph. 
Furthermore, the LPCMCI assumes stationarity, which does not hold when the data has trends or cycles in its values or variance. Trends or cycles can be caused, for example, due to aging or seasonality. 
Another limitation is the assumed absence of selection bias in the data.

While it could be interesting, LPCMCI does not aim to reconstruct the data-generating DAG or its MAG over the observed variables. Also, this work does not evaluate to which extent LPCMCI could do so.
Instead, LPCMCI aims to predict the oracle PAG; thus, the oracle PAG serves as ground truth. Evaluating against the oracle PAG can lead to counter-intuitive results. For example, if LPCMCI predicts the perfect data-generating DAG, the score would be non-perfect if it differs from the oracle PAG. One can see the difference between these two graphs by comparing Figure \ref{fig:predicted}(a) (without the three latent variables and its links) and Figure \ref{fig:predicted}(b).

\begin{figure}[htbp]
    \centering
        \subfigure[Original DAG of the data-generating process.]{
    \includegraphics[width=0.3\linewidth]{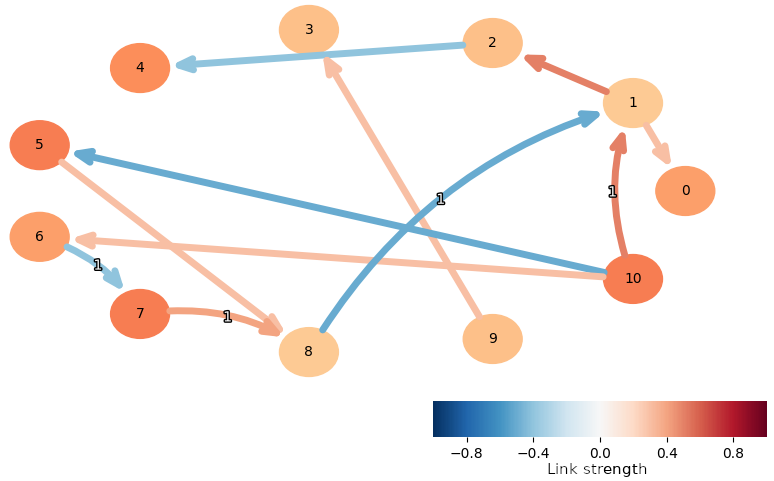}
    }
    \subfigure[Predicted PAG]{
    \includegraphics[width=0.3\linewidth]{figs/5_1.png}
    }
    \subfigure[Predicted PAG without weak links.]{
    \includegraphics[width=0.3\linewidth]{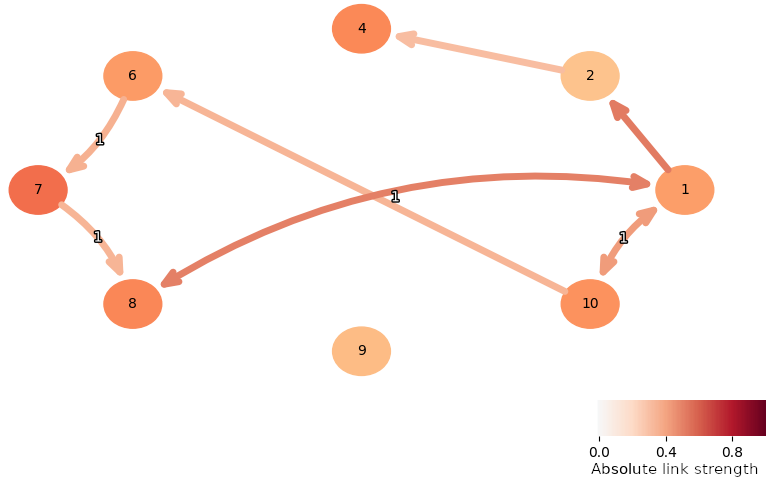}
    }
    \caption{In the case shown here, weak links of the predicted PAG tend to be false positives compared to the original data-generating DAG. Removing weak links removes the false positives without introducing false negatives.}
        \label{fig:removed}
\end{figure}

Figure \ref{fig:removed} shows that the predicted PAG has many false positives with a weak link strength when evaluating against the data-generating DAG. Figure \ref{fig:removed}(c) shows the predicted PAG of the LPCMCI algorithm, where weak links with a strength below 0.10 are removed. In this example of Figure \ref{fig:removed}, we observe that removing weak links decreases false positives without increasing false negatives. In future work, one could investigate this phenomenon in more detail. Moreover, instead of removing weak links at the end, it could be helpful to iteratively lower the significance level $\alpha$ to remove false positives without inducing false negatives.

Another limitation of LPCMCI is that the algorithm only returns the absolute values of the predicted link strength, as one can see in Figure \ref{fig:predicted}. The reason is that it computers the link strength via the p-value. This could be improved by differentiating between positive and negative link strengths using a distance correlation metric (e.g., Pearson correlation coefficient in the linear case).

In this work, we omitted background knowledge. However, using background knowledge is desired as information about (non-)ancestorships may lead to more unambiguous edge orientations and thus increase performance.

\subsection{Conclusion and lessons learned}

To summarize, this work briefly explains the advantages and limitations of today's main methods for observational causal discovery for time series. 
I sampled SCMs that generated multi-dimensional time-series with high auto-correlation and latent variables and applied the LPCMCI algorithm to predict the PAGs that satisfy the observed data's conditional independence constraints and time-order constraints. Lastly, I evaluated the performance of the LPCMCI algorithm and discussed its results. 

The results show that LPCMCI leads to a $harmonic\_score$ of 0.53 while the random baseline scores 0.20.
The significant outperformance of LPCMCI against the random algorithm shows that LPCMCI allows gaining helpful information when the baseline is that we do not know anything. However, one should not rely on the results as the performance of 0.53 is far from the perfect score of 1.00.
As the performance was better than not knowing anything but still far from perfect, we gain predictions but should treat them with high uncertainty. Causal graphs with high uncertainty can still be helpful, for example, when we want to understand the causal relationship of a stochastic process, where observational data is cheap, but conducting experiments are more expensive.
Then we could first apply the LPCMCI algorithm on the cheap observational data to gain knowledge with uncertainty prior to the experiments. When conducting the experiments, we could reduce the number of experiments in cases where a few experiments agree with the results of LPCMCI. 
Similarly, the results of LPCMCI could be used as priors in reinforcement learning to improve efficiency.

During the project, I learned that methods rely on various assumptions and that there is a trade-off in making assumptions. On the one hand, making strong assumptions can make algorithms more powerful. For example, the PCMCI algorithm makes the strong assumption of causal sufficiency, which allows it to predict the more informative DAGs that satisfy the data. In contrast, the LPCMCI does not make this assumption and only predicts the less informative PAG. On the other hand, making strong assumptions entails the risk of violating the assumptions. For example, many algorithms assume there are no latent confounders. Still, in reality, it is often difficult to ensure that it is actually the case, and thus could lead to wrong conclusions.
Furthermore, I learned that predicting and evaluating against the oracle PAG instead of the data-generating DAG can lead to counter-intuitive results. Examples are that a perfect prediction of the data-generating DAG can lead to a non-perfect score or that the performance on contemporaneous links is better than on lagged links, even though the latter allows for time-ordering.

\section{Acknowledgments}
I thank Paul-Christian Buerkner for helpful discussions and suggestions.


\bibliographystyle{IEEEtran}
\bibliography{refs}



\end{document}